\documentclass[3p,times,twocolumn,12pt, procedia]{elsarticle}

\usepackage{ecrc}
\usepackage{color}
\usepackage[hang,flushmargin]{footmisc}
\usepackage{balance}

\volume{Vol. x, No. x}
\firstpage{1}
\journalname{VNU Journal of Science: Comp. Science $\&$ Com. Eng.,}

\runauth{T.T. Nguyen et al.}

\CopyrightLine{2022}{Received x xxxx, Revised x xxxx, Accepted x xxxx}

\usepackage[utf8]{inputenc}
\usepackage[vietnamese,english,icelandic]{babel}
\usepackage{mathtools}
\usepackage{fixltx2e}
\usepackage[T1]{fontenc}
\usepackage{graphicx}
\usepackage{caption}
\usepackage{subcaption}
\usepackage{booktabs}
\usepackage{multirow}
\usepackage[none]{hyphenat}
\usepackage[tracking=true]{microtype}
\usepackage[labelsep=period,justification=centering]{caption}

\renewcommand\footnoterule{\kern-3pt \hrule width 0.65in \kern 2.6pt}
\usepackage{geometry}
\usepackage{hyperref}
\newcommand*\vn{\fontencoding{T5}\selectfont\selectlanguage{vietnamese}}
\begin{document}
\begin{frontmatter}
\title{vieCap4H Challenge 2021: Vietnamese Image Captioning for Healthcare Domain using Swin Transformer and Attention-based LSTM}


\author{Thanh Tin Nguyen$^{1}$}

\author{Long H. Nguyen$^{2}$}

\author{Nhat Truong Pham$^{3,4,}$\corref{label1}}

\author{Liu Tai Nguyen$^{2}$}

\author{Van Huong Do$^{5}$}

\author{Hai Nguyen$^{6}$}

\author{Ngoc Duy Nguyen$^{7}$}

\address{\normalsize $^{1}$ Human Computer Interaction Lab, Sejong University, Seoul, South Korea\\
$^{2}$ Faculty of Information Technology, Ton Duc Thang University, Ho Chi Minh City, Vietnam \\
$^{3}$ Division of Computational Mechatronics, Institute for Computational Science, \\ Ton Duc Thang University, Ho Chi Minh City, Vietnam \\
$^{4}$ Faculty of Electrical and Electronics Engineering, Ton Duc Thang University, \\ Ho Chi Minh City, Vietnam \\
$^{5}$ ASICLAND, Suwon, South Korea \\
$^{6}$ Khoury College of Computer Sciences, Northeastern University, Boston, USA \\
$^{7}$ Institute for Intelligent Systems Research and Innovation, \\ Deakin University, Victoria, Australia\\}

\cortext[label1]{\textls[-30]{~Corresponding author. Email.: phamnhattruong@tdtu.edu.vn}}

\begin{abstract}
\indent

This study presents our approach on the automatic Vietnamese image captioning for healthcare domain in text processing tasks of Vietnamese Language and Speech Processing (VLSP) Challenge 2021, as shown in Figure~\ref{fig:example}. In recent years, image captioning often employs a convolutional neural network-based architecture as an encoder and a long short-term memory (LSTM) as a decoder to generate sentences. These models perform  remarkably well in different datasets. Our proposed model also has an encoder and a decoder, but we instead use a Swin Transformer in the encoder, and a LSTM combined with an attention module in the decoder. The study presents our training experiments and techniques used during the competition. Our model achieves a BLEU4 score of 0.293 on the vietCap4H dataset, and the score is ranked the 3$^{rd}$ place on the private leaderboard. Our code can be found at \url{https://git.io/JDdJm}.
\end{abstract}
\begin{keyword}
image captioning \sep swin transformer \sep encoder-decoder \sep covid-19 \sep health \sep Vietnamese.
\end{keyword}
\end{frontmatter}

\begin{figure}[h]
	\centering
	\includegraphics[width=4.5cm, height=5cm]{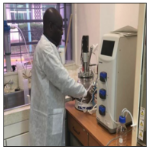}
	\includegraphics[width=7cm, height=2cm]{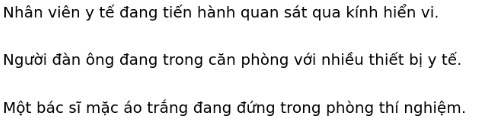}
	\selectlanguage{english}
	\caption{A sample image of the image captioning task. The image is described by three Vietnamese sentences generated by our model.}
	\label{fig:example}
\end{figure}

\section{Introduction}
\label{sec:intro}
Generating meaningful captions for images is recently a challenging topic in artificial intelligence (AI). The task involves both natural language processing (NLP) and computer vision (CV) techniques because it requires the machine to understand an image and translate the understanding into a meaningful caption. Solving the problem leads to several practical applications such as virtual assistants for blind and visually impaired people, conducting visual content indexing and searching. Recently, Vu et al. \cite{vu2020multimodal} also proposed another application of image captioning to generate online personalized reviews via a multimodal approach, such as text, image, and ratings. This helps to understand users' behavior because the generated reviews are based on users' privacy and fairness.

Although the image captioning task has been tackled by a variety of techniques, there are little research on Vietnamese domain. To encourage conducting research on Vietnamese image captioning, \cite{lam2020uit} created a dataset for Vietnamese domain, also serving as a premise for researching on Vietnamese image captioning for healthcare domain.


The vieCap4H Challenge 2021 \cite{vlsp-2021-viecap4h} aims to be a competition for developing machine learning algorithms that use Vietnamese to describe the visual content in healthcare settings, especially images that describe the COVID-19 pandemic. 
Similar to this task, the most recent studies were presented in \cite{vinyals2015show} and \cite{xu2015show} that proposed a network involving a deep Convolutional Neural Network (CNN) as an encoder and a Recurrent Neural Network (RNN) as a decoder. While the encoder is in charge of extracting features from images, the decoder takes these features as input and infers a descriptive caption. The study in \cite{xu2015show} is an extension of the study in \cite{vinyals2015show} by adding an attention mechanism into the decoder part. Inspired by the these studies, we employ a specialized encoder architecture called Swin Transformer \cite{liu2021swin}, and keep the decoder intact as proposed in \cite{xu2015show}.

The rest of the paper is organized as follows. Section 2 presents related studies that we have investigated. Section 3 describes our proposed method and the different techniques used during the competition. Analytical results are presented in Section 4, and a detailed discussion is presented in Section 5. Finally, Section 6 concludes the paper and outlines potential future directions.

\section{Related Work}
\label{sec:related_work}

Our approach was initially inspired by the early works in \cite{vinyals2015show} and \cite{xu2015show} for image captioning task. Additionally, we also reference the code from \cite{code}. Recently introduced ideas of using CNNs and RNNs can be found in \cite{vinyals2015show} and \cite{xu2015show}. Some proposed bottom-up methods are mentioned in \cite{pedersoli2017areas}, \cite{herdade2019image}, or \cite{anderson2018bottom}. These bottom-up methods all require to incorporate an object detection module to extract object proposals, then give them to the encoder.

Although CNN has a profound impact on CV, Transformer, first introduced in the paper ``Attention is all you need" \cite{vaswani2017attention}, has progressively replaced CNN in this field. There are plenty of variants of Transformer, but the two most well-known used for extracting visual contents are the Vision Transformer \cite{dosovitskiy2020image} and the Swin Transformer~\cite{liu2021swin} architecture.

Due to the time and resource limitation, our team only improved the encoder side by using the Swin Transformer.

\selectlanguage{english}
\section{Methodology}
\label{sec:method}
\subsection{Attempted Techniques}
Our attempts in this competition consist of the following:
\begin{itemize}
\item Changing the architecture of the encoder by using different CNN architecture.
\item In the decoder, RNN and Transformer were adopted to extract a caption for an image.
\item Tuning the image size, the learning rate, and the learning rate scheduler.
\item Scrubbing the ground-truth captions, as well as attempting to use a pre-trained embedding for text.
\item Augmenting images and text.
\item Using noise injection.
\item Experimenting beam search with different beam widths.
\item Using cross validation.
\item Attempting ensemble models.
\end{itemize}

We start from the model proposed by \cite{vinyals2015show}. Moreover, we also test the idea of predicting the English captions of images and then translating captions from English to Vietnamese. However, this approach failed because translation cannot capture the semantic and relational information of a sentence.

\subsection{Pre-processing}
The data in the challenge consists of two training datasets containing 8,032 images and a validation dataset containing 1,002 images. Both datasets are used in the public phase of the competition. Additionally, there is another testing dataset containing 1,032 images.

In the public training dataset, there are a few images with defective captions, such as repeated or redundant characters in a word, two consecutive words are stick together, or English instead of Vietnamese captions. Due to these erroneous labels, the initial performance of the model is poor and unstable. After fixing the errors, we also convert all of the captions to lowercase and remove all existing punctuation and numbers.

\begin{figure*}
	\centering
	\includegraphics[width=0.97\textwidth]{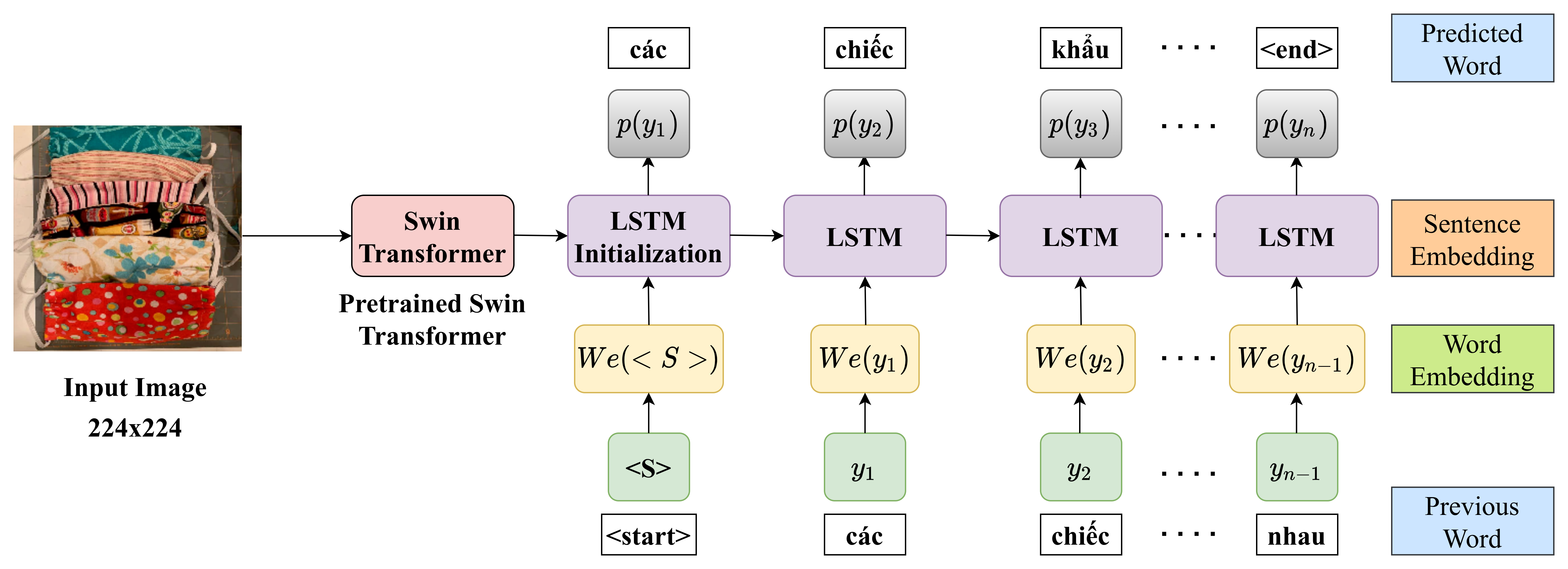}
	\caption{Our proposed model has two main components: (1) The encoder uses Swin Transformer pre-trained on the ImageNet \cite{krizhevsky2012imagenet} dataset; (2) The decoder consists several layers of LSTM.}
	\label{fig:model}
\end{figure*}

\subsection{Model Architecture}
Our proposed model as shown in Figure \ref{fig:model} is an end-to-end integration of a Swin Transformer encoder and attention-based LSTM decoder. These architectures were chosen because, firstly, the Swin Transformer presents a structure in which an image is divided into multiple patches and each patch is divided into multiple windows to perform self-attention within the window. Also, it allows a hierarchical structure resulting in a good performance in several tasks such as object detection, and image segmentation. Secondly, the LSTM is well-known and straightforward for capturing the semantic meaning of natural languages.

\begin{figure}[h]
	\centering
	\includegraphics[width=\linewidth]{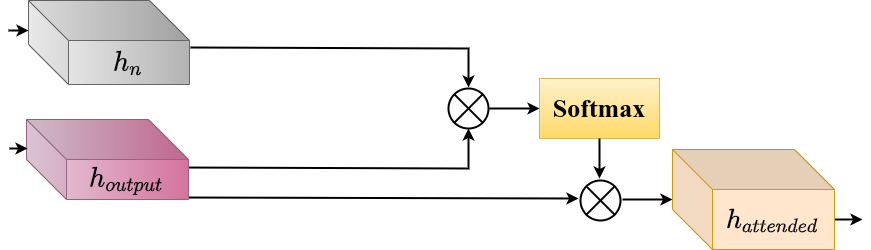}
	\caption{Our attention module:  The hidden state $h_{n}$ is multiplied with the encoded state $h_{output}$, then pass the output through a softmax function to create a weighted distribution over the $h_{output}$. Finally, $h_{attended}$ is computed by multiplying the weighted distribution with the $h_{output}$ to highlight salient regions.}
	\label{fig:sa}
\end{figure}

The model takes in an image and outputs the corresponding caption as follows. Firstly, the pre-processed image with the size of 224$\times$224 is fed into the pre-trained Swin Transformer encoder to extract visual features. Then these extracted visual features are used as the inputs of the attention-based LSTM decoder to generate the caption. This attention step is called late fusion because it aggregates two features extracted from two different architectures. More concretely, the attention module calculates how much attention to a specific hidden vector is given to an output hidden vector. In this case, $h_{T}$ is the last hidden vector of the LSTM,  $h_{output}$ is the output of the encoder which is Swin Transformer, after computing attention, it will output $h_{attended}$ state, this is the representation vector for decoding step. The self-attention can be seen in Fig. \ref{fig:sa}. Finally, post-processing is applied using beam search to improve the accuracy of the proposed model.

\subsection{Noise Injection}
In the field of image captioning, the accuracy of prediction for the next character can be relatively high if we can ensure that the previous sequence is predicted correctly. However, if a character is mispredicted, then the prediction failure rate will gradually increase.

To tackle the above issue, we use noise injection, which is done by randomly replacing ground truth characters with other characters during training, and new sentences will be assigned fake labels called $fake$. Then, using the modified sentences as the input for the decoder will force it to correctly predict the next character on the basis that the previous one was wrong. Specifically, there will be two losses, the cross-entropy loss between the ground truth and the prediction, and the same loss however between the $fake$ and their predictions. In the final loss, we combine the two losses with a weighting $\beta = 0.1$ for the second loss
\begin{equation}
\mathcal{L}_{CE} = -\sum_{i=1}^{N}y^{(i)}*\log{\hat{y}^{(i)}} - \beta\sum_{i=1}^{N}y^{(i)}*\log{\hat{y}^{(i)}_{fake}},
\end{equation}
where $\mathcal{L}_{CE}$ is the Cross Entropy loss, $y^{(i)}$ and $\hat{y}^{(i)}$ denote a label and its predicted probability, respectively. $N$ represents the total number of classes.

\subsection{Beam Search}
Beam search is a well-known search algorithm adopted in several problems such as Image Captioning, Sequence to Sequence. Normally, at test time, in order to generate the output text, greedy search is employed to get the maximum token at each time $t$. This algorithm can not assure to extract the best sequence, because of that, beam search comes into rescue to keep a set of $k$ (beam size) best tokens at each time step. This results in a best approximately sentence.
We tried beam search sizes from 1 to 10, and the best BLEU4 score is achieved with a beam size of 2. Thus, we chose 2 as the beam search size.

\subsection{Training Configuration}
All experiments were trained on a single Titan Xp GPU. The batch size is 16, the input image size is 224$\times$224, the learning rate of the encoder is 1e-4, that of the decoder is 4e-4, the Adam optimizer is used in this model with a weight decay of 1e-6. Moreover, the Cosine Annealing Warm Restarts~\cite{loshchilov2016sgdr} scheduler is used for scheduling the learning rate. In addition, the model uses $k$-fold cross validation with $k=4$. We also use common augmentation techniques such as \textit{HorizontalFlip, RandomCrop} with probability of 0.5, and \textit{Normalize} with mean and std are (0.485, 0.456, 0.406), (0.229, 0.224, 0.225), respectively. Finally, our model uses a cross entropy as the loss function.

\section{Analysis}
\label{sec:exp}
\subsection{Results}
\paragraph{BLEU4 Scores:} The model was tested on a private test set which contains 1,032 images. In this competition, the BLEU4 score is used as the metric to evaluate models. Our model achieves the BLEU4 score of 0.293 and is ranked the 3$^{rd}$ on the private leaderboard. In terms of the public test dataset, which contains 1,002 images, the highest BLEU4 score evaluated on this dataset is 0.302. The results during the public phase are summarized in Table \ref{tab:compare_table}.

\begin{table*}[h!]
	\centering
	\caption{  The table shows the performance of our approach by using a CNN-based model to the Transformer model evaluated on the public dataset. By using additional techniques such as beam search, noise injection, and augmentation, we achieved the highest score on the public test dataset, which is 0.302.}\label{tab:compare_table}
	\begin{tabular}{ |p{3cm}|p{3cm}|p{4cm}|p{1.2cm}|  }
		\hline
		\textbf{Encoder} & \textbf{Decoder} & \textbf{Additional Methods} & \textbf{BLEU4}\\
		\hline
		Resnet101 & LSTM+Attention  & Beam search & 0.253\\
		Efficientnetv2 & LSTM+Attention  & Beam search & 0.266\\
		Vision Transformer & LSTM+Attention  & Beam search & 0.277\\
		Swin Transformer & LSTM+Attention  & Beam search & 0.292\\
		Swin Transformer & LSTM+Attention  & {Beam search + Noise Injection + Augmentation} & 0.302\\
		\hline
	\end{tabular}
	
\end{table*}

\paragraph{Computational processing times and resources:} In this study, the proposed method is implemented with one GPU Titan Xp. Besides, our proposed method takes about 4 hours to train with 4-fold cross validation.

\subsection{Visualization}
\begin{figure*}[!h]
	
	\centering
	\includegraphics[width=0.35\linewidth]{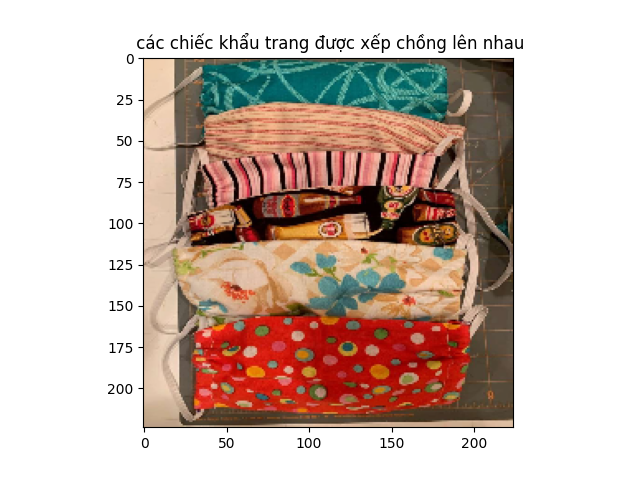}
	\includegraphics[width=0.9\linewidth]{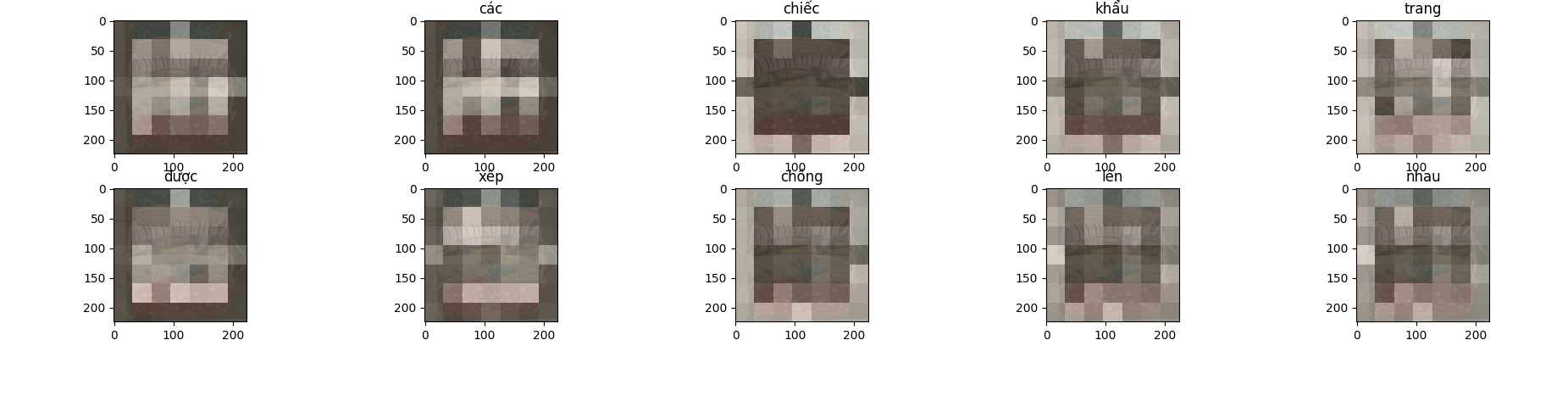}
	\\ \small (a) Attention visualization of \vn `Các chiếc khẩu trang được xếp chồng lên nhau' sample. \\

	
	
	
	\centering
	\includegraphics[width=0.35\linewidth]{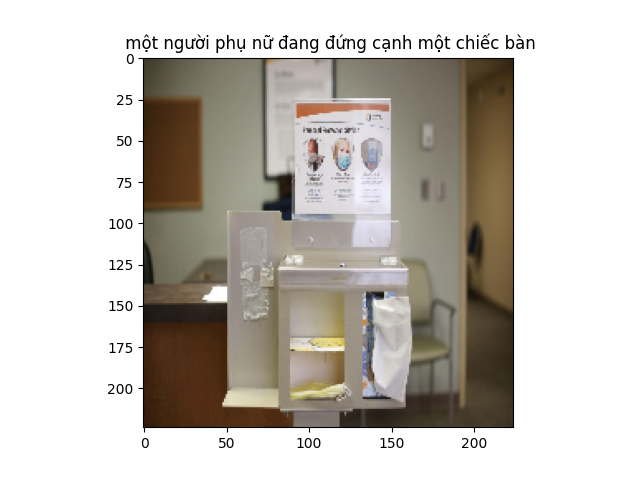}
	\includegraphics[width=0.9\linewidth]{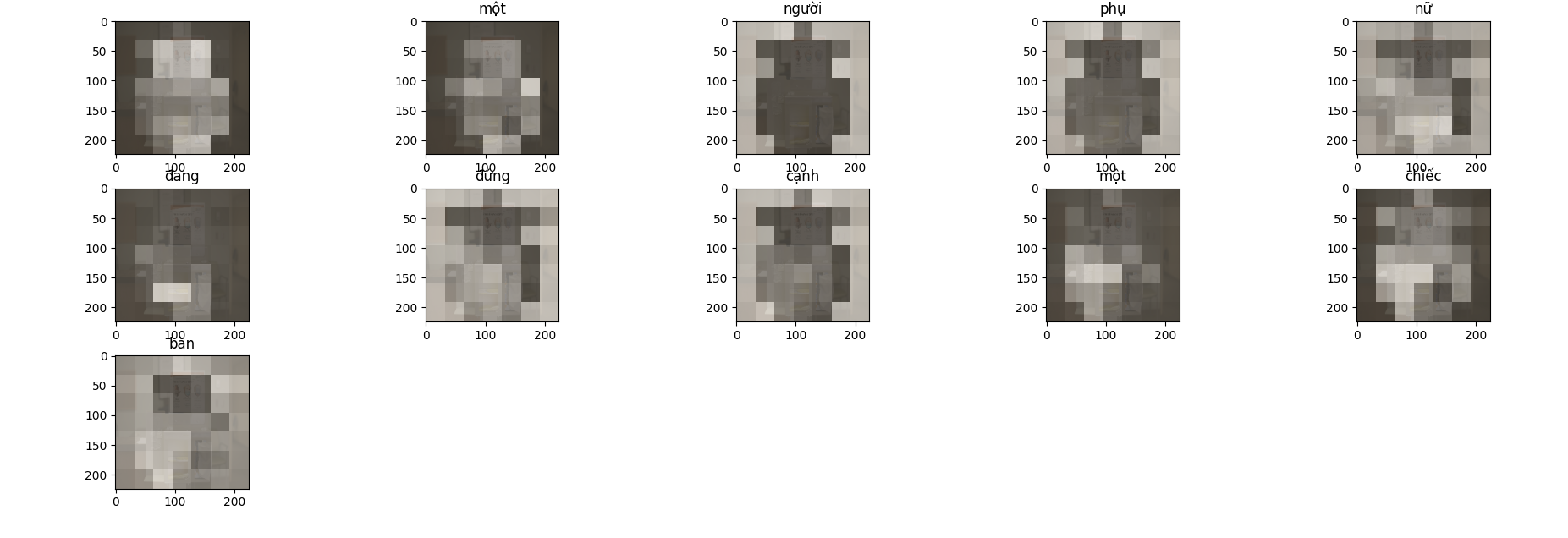}
	\\ \small (b) Attention visualization of \vn `Một người phụ nữ đang đứng cạnh một chiếc bàn' sample.\\
	
    \selectlanguage{english}
	\caption{Attention visualization of test samples. In Figure \ref{fig:att}a, the words \vn `khẩu'~and `trang'~are attending in the middle of the image where exactly the object \vn `khẩu trang'~is located. Figure \ref{fig:att}b describes a mistaken caption of its corresponding image.}
	\label{fig:att}
	
\end{figure*}

Visualizing attention maps is to explain how the model learns to concentrate on different parts of an image with its corresponding words. These attention images are shown in Figure \ref{fig:att}. For example, the caption of the image in Figure \ref{fig:att}a is \vn `Các chiếc khẩu trang được xếp chồng lên nhau' in which the words \vn `khẩu'~and `trang'~are attending in the middle of the image where exactly the object \vn `khẩu trang'~is located. Figure \ref{fig:att}b describes a mistaken caption of its corresponding image. The caption of the image in Figure \ref{fig:att}b is \vn `Một người phụ nữ đang đứng cạnh một chiếc bàn', however, there is no \vn `một người phụ nữ' (a woman) in the image.

\subsection{Ablation tests}
We have performed several ablation tests before perfecting the final model. Table \ref{tab:ab_table} shows our analysis of the contribution of each process in the model including: without pre-processing data, without beam search, without noise injection, and the final model.

\begin{table}[h!]
	\centering
	\selectlanguage{english}
	\caption{The table shows the score of our final model and when removing important parts in the model evaluated on the public dataset.}\label{tab:ab_table}
	\begin{tabular}{ |p{5cm}|p{1.7cm}|  }
		\hline
		\textbf{Model} & \textbf{BLEU4}\\
		\hline
		Without pre-processing data & 0.273\\
		Without beam search & 0.286\\
		Without noise injection & 0.293\\
		\textbf{Our final model} & \textbf{0.302}\\
		\hline
	\end{tabular}

\end{table}

\subsection{Discussion}
We found that the provided data has some minor errors such as false captions, redundant words, and typos, and after fixing all these errors, the accuracy increased a little bit. Beam search is a must for efficient decoding, in fact, it helps us boost the performance. Finally, with the noise injection in the loss function, the model can predict the next character more properly. In conclusion, following the encoder-decoder manner, we adopted two well-known architectures which are Swin Transformer and LSTM, and the result has demonstrated that our proposed model is a powerful tool for image captioning.

\section{Lessons Learned}
We have tried a variety of techniques, some perform well, and some do not. In this section, we describe different techniques that are used during the competition.
\subsection{Working Techniques}
Firstly, we examine the training data and apply suitable adjustments to defective captions. During the tuning process, we use a grid search of different learning rates and image sizes.

Swin Transformer~\cite{liu2021swin} is also a key factor that outperforms most CNN-based models, including Efficientnet~\cite{tan2019efficientnet}, Efficientnet v2~\cite{tan2021efficientnetv2}, Restnet 50~\cite{he2016deep}, and Resnet 101~\cite{he2016deep} concerning the accuracy, training time, and memory capacity. It also outperforms the Vision Transformer~\cite{dosovitskiy2020image}, which is another variant of a Transformer.

Augmentation is a promising method to ensure the model capacitates with different transformations of the images. In this technique, we use a \textit{Resize} factor of 224 for each dimension, a \textit{HorizontalFlip} factor of 0.5, and a \textit{RandomCrop} factor of 0.5. Finally, a \textit{Normalization} factor is used with a mean of (0.485, 0.456, 0.406) and a standard deviation of (0.229, 0.224, 0.225).

Noise injection is a regularization technique used to randomly replace ground truth characters with other characters during training to create new sentences. This technique helps to improve the prediction of the next character while the previous one is imperfect.

Beam search is a technique that can seek the best potential sentence while decoding the features. When cross-validating, we run our model on four different subsets of the data to achieve different outcome measures. Although it can take a long time, we can achieve a better result.

\subsection{Non-working Techniques}
We conduct the performance evaluation by using two different image sizes: 224$\times$224 and 384$\times$384. It is surprising that an image size of 224$\times$224 provides a better result compared with an image size of 384$\times$384.

Various CNN-based models such as Efficientnet \cite{tan2019efficientnet}, Efficientnetv2 \cite{tan2021efficientnetv2}, Resnet 50 \cite{he2016deep}, Resnet 101 \cite{he2016deep}, or Vision Transformer \cite{dosovitskiy2020image} (ViT) have been evaluated as the encoder. They consume a higher memory and have low performance in terms of accuracy and training time. 

Although pre-trained word embedding models, e.g., \cite{bojanowski2017enriching} and \cite{nguyen2020pilot}, have trained on large corpus in Vietnamese, we failed to incorporate it into our model due to a small embedding size of the pre-trained ones.

We also employed the Transformer as the decoder, but the achieved accuracy was not as good as that achieved by the LSTM. Moreover, we also failed at implementing beam search for the Transformer decoder due to its complicated architecture.

Finally, ensemble is a regularization technique that helps boost overall accuracy, but unfortunately, we failed to implement it because of lack of time.

\subsection{Unique domain/data specific insights that you have uncovered}
Noise injection, beam search, and cross validation are unique techniques that helps to improve the accuracy. Moreover, carefully screening the captions also helps the model learn better. Although these are not new and novel to somebody, but these still gave us surprises, and we have learnt a lot from it.

\selectlanguage{english}
\section{Conclusion}
\label{sec:con}
This paper describes our approach in the automatic Vietnamese image captioning for healthcare domain in text processing tasks of the VLSP Challenge 2021. The proposed model employs an encoder-decoder architecture, and experimental results show a potentially useful network for tackling the problem. Furthermore, techniques such as noise injection, beam search, and cross validation help boost the algorithm's performance during the competition. Finally, cleaning and pre-processing the data improves the performance of the algorithm.

One potential direction to further improve the performance of the decoder such as using a Transformer-based decoder. Additionally, using beam search while decoding, there are repeated words. Fixing this problem would potentially increase the performance. 



\balance
\bibliographystyle{elsarticle-num}


\begin{thebibliography}{10}
	\expandafter\ifx\csname url\endcsname\relax
	\def\url#1{\texttt{#1}}\fi
	\expandafter\ifx\csname urlprefix\endcsname\relax\def\urlprefix{URL }\fi
	\expandafter\ifx\csname href\endcsname\relax
	\def\href#1#2{#2} \def\path#1{#1}\fi
	
	\bibitem{vu2020multimodal}
	X.-S. Vu, T.-S. Nguyen, D.-T. Le, L.~Jiang, Multimodal review generation with
	privacy and fairness awareness, in: 28th International Conference on
	Computational Linguistics (COLING), Barcelona, Spain (Online), December 8-13,
	2020., International Committee on Computational LinguisticsInternational
	Committee on Computational Linguistics, 2020, pp. 414--425.
	
	\bibitem{lam2020uit}
	Q.~H. Lam, Q.~D. Le, V.~K. Nguyen, N.~L.-T. Nguyen, Uit-viic: A dataset for the
	first evaluation on vietnamese image captioning, in: International Conference
	on Computational Collective Intelligence, Springer, 2020, pp. 730--742.
	
	\bibitem{vlsp-2021-viecap4h}
	T.~M. Le, L.~H. Dang, T.-S. Nguyen, T.~M.~H. Nguyen, X.-S. Vu, Vlsp 2021 -
	viecap4h challenge: Automatic image caption generation for healthcare domain
	in vietnamese, in: Proceedings of the 8th International Workshop on
	Vietnamese Language and Speech Processing, VNU Journal of Science: Computer
	Science and Communication Engineering, HCM, Vietnam, 2021.
	
	\bibitem{vinyals2015show}
	O.~Vinyals, A.~Toshev, S.~Bengio, D.~Erhan, Show and tell: A neural image
	caption generator, in: Proceedings of the IEEE conference on computer vision
	and pattern recognition, 2015, pp. 3156--3164.
	
	\bibitem{xu2015show}
	K.~Xu, J.~Ba, R.~Kiros, K.~Cho, A.~Courville, R.~Salakhudinov, R.~Zemel,
	Y.~Bengio, Show, attend and tell: Neural image caption generation with visual
	attention, in: International conference on machine learning, PMLR, 2015, pp.
	2048--2057.
	
	\bibitem{liu2021swin}
	Z.~Liu, Y.~Lin, Y.~Cao, H.~Hu, Y.~Wei, Z.~Zhang, S.~Lin, B.~Guo, Swin
	transformer: Hierarchical vision transformer using shifted windows, arXiv
	preprint arXiv:2103.14030.
	
	\bibitem{code}
	S.~Vinodababu,
	https://github.com/sgrvinod/a-p-\\ytorch-tutorial-to-image-captioning.
	
	\bibitem{pedersoli2017areas}
	M.~Pedersoli, T.~Lucas, C.~Schmid, J.~Verbeek, Areas of attention for image
	captioning, in: Proceedings of the IEEE international conference on computer
	vision, 2017, pp. 1242--1250.
	
	\bibitem{herdade2019image}
	S.~Herdade, A.~Kappeler, K.~Boakye, J.~Soares, Image captioning: Transforming
	objects into words, arXiv preprint arXiv:1906.05963.
	
	\bibitem{anderson2018bottom}
	P.~Anderson, X.~He, C.~Buehler, D.~Teney, M.~Johnson, S.~Gould, L.~Zhang,
	Bottom-up and top-down attention for image captioning and visual question
	answering, in: Proceedings of the IEEE conference on computer vision and
	pattern recognition, 2018, pp. 6077--6086.
	
	\bibitem{vaswani2017attention}
	A.~Vaswani, N.~Shazeer, N.~Parmar, J.~Uszkoreit, L.~Jones, A.~N. Gomez,
	{\L}.~Kaiser, I.~Polosukhin, Attention is all you need, in: Advances in
	neural information processing systems, 2017, pp. 5998--6008.
	
	\bibitem{dosovitskiy2020image}
	A.~Dosovitskiy, L.~Beyer, A.~Kolesnikov, D.~Weissenborn, X.~Zhai,
	T.~Unterthiner, M.~Dehghani, M.~Minderer, G.~Heigold, S.~Gelly, et~al., An
	image is worth 16x16 words: Transformers for image recognition at scale,
	arXiv preprint arXiv:2010.11929.
	
	\bibitem{krizhevsky2012imagenet}
	A.~Krizhevsky, I.~Sutskever, G.~E. Hinton, Imagenet classification with deep
	convolutional neural networks, Advances in neural information processing
	systems 25 (2012) 1097--1105.
	
	\bibitem{loshchilov2016sgdr}
	I.~Loshchilov, F.~Hutter, Sgdr: Stochastic gradient descent with warm restarts,
	arXiv preprint arXiv:1608.03983.
	
	\bibitem{tan2019efficientnet}
	M.~Tan, Q.~Le, Efficientnet: Rethinking model scaling for convolutional neural
	networks, in: International Conference on Machine Learning, PMLR, 2019, pp.
	6105--6114.
	
	\bibitem{tan2021efficientnetv2}
	M.~Tan, Q.~V. Le, Efficientnetv2: Smaller models and faster training, arXiv
	preprint arXiv:2104.00298.
	
	\bibitem{he2016deep}
	K.~He, X.~Zhang, S.~Ren, J.~Sun, Deep residual learning for image recognition,
	in: Proceedings of the IEEE conference on computer vision and pattern
	recognition, 2016, pp. 770--778.
	
	\bibitem{bojanowski2017enriching}
	P.~Bojanowski, E.~Grave, A.~Joulin, T.~Mikolov, Enriching word vectors with
	subword information, Transactions of the Association for Computational
	Linguistics 5 (2017) 135--146.
	
	\bibitem{nguyen2020pilot}
	A.~T. Nguyen, M.~H. Dao, D.~Q. Nguyen, A pilot study of text-to-sql semantic
	parsing for vietnamese, arXiv preprint arXiv:2010.01891.
	
\end{thebibliography}

\end{document}